\documentclass[letterpaper, 10 pt, conference]{ieeeconf}  % Comment this line out if you need a4paper

\IEEEoverridecommandlockouts                              % This command is only needed if 
\usepackage[linesnumbered,lined,ruled,commentsnumbered]{algorithm2e}
\usepackage{algpseudocode} 
\usepackage{graphics} 
\usepackage{epsfig}
\usepackage{multirow}
\usepackage{numprint} %rounding numbers
\usepackage{caption}
\usepackage{amsmath,amssymb,amsfonts}

\usepackage{atbegshi}
\usepackage[absolute,overlay]{textpos}
\usepackage{rotating}
\usepackage[english]{babel}
\usepackage{blindtext}

\usepackage{tikz,xcolor,hyperref}
\definecolor{lime}{HTML}{A6CE39}
\DeclareRobustCommand{\orcidicon}{
	\begin{tikzpicture}
	\draw[lime, fill=lime] (0,0) 
	circle [radius=0.16] 
	node[white] {{\fontfamily{qag}\selectfont \tiny ID}};
	\draw[white, fill=white] (-0.0625,0.095) 
	circle [radius=0.007];
	\end{tikzpicture}
	\hspace{-2mm}
}
\foreach \x in {A, ..., Z}{\expandafter\xdef\csname orcid\x\endcsname{\noexpand\href{https://orcid.org/\csname orcidauthor\x\endcsname}
			{\noexpand\orcidicon}}
}
\title{\LARGE \bf
Integrating Naturalistic Insights in Objective Multi-Vehicle Safety Framework
}

\author{Enrico Del Re\orcidE{}, \emph{Graduate Student Member, IEEE} Amirhesam Aghanouri\orcidA{}\\ Cristina Olaverri-Monreal\orcidC{} \emph{Senior Member, IEEE}% <-this % stops a space
\thanks{
        Johannes Kepler University Linz, Austria; Department Intelligent Transport Systems \texttt{\{enrico.del\_re, amirhesam.aghanouri, cristina.olaverri-monreal\}@jku.at}
        }
    }

\begin{document}

\maketitle
\thispagestyle{empty}
\pagestyle{empty}

%%%%%%%%%%%%%%%%%%%%%%%%%%%%%%%%%%%%%%%%%%%%%%%%%%%%%%%%%%%%%%%%%%%%%%%%%%%%%%%%
\begin{abstract}
As autonomous vehicle technology advances, the precise assessment of safety in complex traffic scenarios becomes crucial, especially in mixed-vehicle environments where human perception of safety must be taken into account. This paper presents a framework designed for assessing traffic safety in multi-vehicle situations, facilitating the simultaneous utilization of diverse objective safety metrics. Additionally, it allows the integration of subjective perception of safety by adjusting model parameters. The framework was applied to evaluate various model configurations in car-following scenarios on a highway, utilizing naturalistic driving datasets. The evaluation of the model showed an outstanding performance, particularly when integrating multiple objective safety measures. Furthermore, the performance was significantly enhanced when considering all surrounding vehicles.

\end{abstract}

\begin{textblock*}{18.15cm}(1.55cm,26cm) % Adjust width and position as needed
\begin{minipage}{17.8cm}
     \vspace{0.1cm} % Vertical space within the minipage
     {\footnotesize\copyright 2024 IEEE. Personal use of this material is permitted. Permission from IEEE must be obtained for all other uses, in any current or future media, including reprinting/republishing this material for advertising or promotional purposes, creating new collective works, for resale or redistribution to servers or lists, or reuse of any copyrighted component of this work in other works. DOI: 10.1109/ITSC58415.2024.10920258}
\end{minipage}
\end{textblock*}

%%%%%%%%%%%%%%%%%%%%%%%%%%%%%%%%%%%%%%%%%%%%%%%%%%%%%%%%%%%%%%%%%%%%%%%%%%%%%%%%
\section{INTRODUCTION}
\label{sec:introduction}
The progression of Advanced Driver Assistance Systems (ADAS) has significantly enhanced autonomous vehicle technology, playing a crucial role in reducing traffic accidents \cite{ADAS_impact} and aligning with the EU's ``Vision-Zero'' initiative \cite{european2019eu}. Certain systems, like the Lane-Keeping Assistant (LKA), require compliance to complex safety regulations such as UN 157 \cite{UN157} to ensure robust safety standards.

In this context, Surrogate Safety Measures (SSMs) become relevant as indirect indicators, providing a method to evaluate safety-related aspects in traffic and transportation, particularly in the absence of actual crash data. These measures are often deduced from observable driving parameters and behaviors, that are influenced by objective driving parameters that are measurable and quantifiable (e.g. speed, distance traveled, acceleration, deceleration, vehicle position, etc.) without subjective interpretation. For example, headway (time or distance between a vehicle and the one in front of it) represents a driving parameter associated with surrogate safety measures. 

SSMs are widely employed in safety regulations. However, they might show limitations when characterizing specific traffic situations. For example, a vehicle tailgating another might be deemed safe according to certain SSMs but critical according to others. 
This discrepancy arises when one SSM relies solely on the absolute distance between vehicles as a safety indicator and another SSM considers additional factors, such as relative speed, time-to-collision, or acceleration patterns.

Moreover, SSMs are typically designed for interactions between two vehicles, thereby restricting their applicability to more complex traffic scenarios.

Furthermore, although SSMs are correlated with accident risk, discrepancies can emerge between their safety assessments and human acceleration patterns when applied to the analysis of natural driving behavior \cite{DSSM}. As an example, drivers may not respond immediately to a vehicle cutting in front of them after overtaking, as the situation is anticipated to stabilize quickly.
This discrepancy between SSMs and the human perception of safety has been thoroughly documented in two prior publications, \cite{10186747} and \cite{9987124}, with a particular focus on highway scenarios, in the context of lane changes between two vehicles.

Building upon the results from the aforementioned findings, the application or assessment of SSMs can involve subjective factors related to human perceptions and interpretations. Therefore, we contribute to the state of the art with a new comprehensive safety model framework that integrates those subjective factors, which reflect the human perception of risk (e.g. relative position of other vehicles), with multi-vehicle application of SSMs for a more detailed assessment of traffic safety that better aligns with human judgement and driving behavior. Bridging the gap between objective and subjective risk assessment is essential not only for developing human-like AVs, but also for designing robust crash avoidance systems to mitigate risks, for example in scenarios associated with hydrogen-based vehicle and electric vehicle battery incidents.

We evaluated the safety model framework in car following scenarios on the highway, utilizing naturalistic driving datasets from both a drone viewpoint (HighD dataset \cite{highD}) and an ego vehicle standpoint (IAMCV dataset \cite{icmcv, certad2023}). The pertinent statistical tests were then conducted to assess the effects of using a combination of SSMs that include all surrounding vehicles. Additionally, adjustments were made based on relative positions (e.g., leading, following, and driving in parallel) to evaluate the model's improved alignment with naturalistic driving data.

The following section provides an overview of current safety models, emphasizing the key differences from our proposed model. In Section \ref{sec:Scenario}, the traffic scenario and datasets utilized for model evaluation are introduced. Section \ref{sec:Methodology} presents the structure of the safety model and Section \ref{evaluation} outlines the model evaluation. The results are presented in Section \ref{sec:Results}. Lastly, Section \ref{sec:Conclusion} summarizes the key findings and outlines potential future directions for this research.

\section{Related Work}
\label{sec:RelatedWork}
While SSMs have been extensively studied, as evidenced by reviews in \cite{SSM} and \cite{SSM_2023}, the majority of research efforts have been directed toward enhancing their accuracy in two-vehicle scenarios. Improvements, such as modifications to existing SSMs \cite{MTTC} or the development of new ones with reduced dependency on fixed thresholds, as seen in \cite{fuzzy}, have reinforced their reliability. In addition, some applications emerged through the combination of different SSMs \cite{SSAM} including the formulation of new safety measures \cite{Combined_SSM}. However, challenges posed by multi-vehicle scenarios have not been sufficiently addressed.

Research on the perception of traffic safety has seldom been performed around objective parameters. Questionnaires exploring the impact of various factors on traffic safety — such as driving under the influence of alcohol or exceeding speed limits \cite{ESRA} — offer valuable insights, yet they pose challenges when integrating into models depending exclusively on objective safety metrics, especially without in-vehicle monitoring.

Studies investigating perceptions of Time-To-Collision \cite{TTC_perception} or distances using convex mirrors \cite{mirror} are more aligned with objective parameters. Their results contributed to the position dependent weights used within this paper, though more precise modeling requires additional data, like mirror curvature and angle \cite{mirror_adjustment}, which is exceedingly difficult to acquire.

The safety field theory consistently incorporates the assessment of multiple vehicles and infrastructure \cite{safety_field_infra} treating other road users as entities that engage in mutual interactions~\cite{physics_field}. While calculating a force acting on the ego vehicle and labeling it as a safety measure is possible, the safety field theory predominantly prioritizes path planning, presenting a significant challenge in effectively modeling the perception of safety by human drivers.

Modeling the subjective perception of safety presents a notable challenge due to a scarcity of labeled data — specifically, data encompassing both objective and subjective measures.  In \cite{SSM_model_india} such a dataset was created by conducting assessments of traffic conflicts by trained observers. This method enabled the identification of objective factors predominantly influencing individuals' safety perceptions, specifically those affecting both the proximity and severity of potential collisions. However, it is crucial to note that the scenarios analyzed were limited to interactions involving only two vehicles, and the labor-intensive nature of labeling interactions makes this approach less practical for more complex scenarios.

To address this challenge, the safety model framework presented in this paper utilized SSMs as the base estimator, for an ease of calculation, association with accidents, and interpretability. Additionally, the model incorporated a safety field-like consideration, taking into account the 2-dimensional positions of other vehicles.

\section{Traffic Scenario and Dataset description}
\label{sec:Scenario}
\subsection{Traffic scenario}
We focused on traffic scenarios on the highway, where the ego vehicle maintains a relatively constant trajectory without executing lane changes. Consequently, the sole response by the human driver of the ego vehicle to variations in traffic safety was reflected in changes in acceleration. 
Therefore, the association between the safety model and human perception of safety could be examined through a bivariate correlation test involving the jerk and gradient of the safety risk.

\begin{figure}
    \centering
    \includegraphics[scale=0.33]{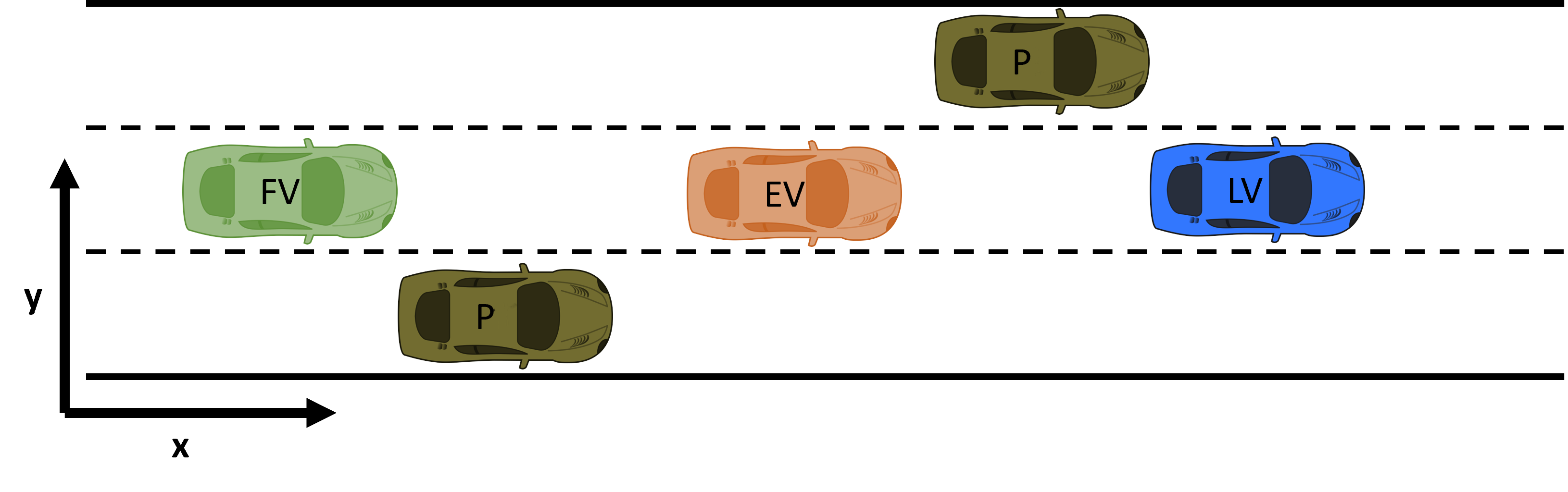}
    \caption{Illustration of the safety analysis of the examined traffic scenario. Each ego vehicle (EV) is evaluated along with its leading vehicle (LV), following vehicle (FV), and parallel-moving vehicles (P).}
    \label{fig:traffic_scenario}
\end{figure}

The trajectories of the surrounding vehicles might involve lane changes or other maneuvers. We examined both the leading (LV) and following (FV) vehicles, along with vehicles moving in parallel in the adjacent lane to the ego vehicle, both in a leading (PL) and following (PF) position, as surrounding vehicles, as illustrated in Figure \ref{fig:traffic_scenario}. Traffic scenarios involving lane changes by trucks were excluded from consideration.

\subsection{Datasets}

The model was tested using trajectories from the HighD dataset \cite{highD}. This dataset included car and truck trajectories on German highways, recorded by a drone over a total duration of 16.5 hours. Only the first 57 recordings were utilized, since the remaining 3 involve highway ramps which could heavily influence the driving behavior. 

Additionally, we demonstrated how the model could be applied from the ego-vehicle perspective using a highway scene from the IAMCV dataset \cite{icmcv}, \cite{certad2023}. This extensive dataset recorded over 14 hours of data from the viewpoint of an ego vehicle navigating various road environments, including highways, roundabouts, and intersections. For our analysis, we selected and labeled a scene from the A3 Federal Motorway in Germany, as it provided a good long term overview on the traffic situation for the ego vehicle.

\section{Safety Model}
\label{sec:Methodology}

\subsection{Metrics}

To assess safety between two vehicles in a comprehensive way,   in scenarios involving trajectory intersections and car-following situations, we incorporated the following metrics into our safety model.  
\begin{itemize}
\item Post Encroachment Time (PET): Time duration between the first vehicle leaving a specific point of intersection and the subsequent arrival of the second vehicle at the same point. In car-following scenarios, it is identical to Time Headway (TH).
%\item Time Headway (TH): duration for the rear of one vehicle to traverse a specific point on the road, followed by the front of the subsequent vehicle passing the same reference point.
\item Time-To-Collision (TTC): Time for a vehicle to reach a potential collision point if it maintains its current speed and trajectory. 
\item Minimum Deceleration Required to Avoid Crash (DRAC): Lowest deceleration rate that a vehicle must achieve in order to prevent a collision with another vehicle.

\end{itemize}

TTC and DRAC were selected as SSMs due to their suitability for car-following scenarios. While TH could smoothly apply to car-following situations, we opted to use PET instead.  This allows us to evaluate safety implications related to the parallel-moving vehicles in our scenario.

In cases where the trajectories of two vehicles intersected, PET is defined by the difference between the time until the first vehicle leaves the crossing point of the intersection ($t_1$) and the time ($t_2$) when the second vehicle arrives at the same point, as formally defined in equation~\ref{eq1},

\begin{equation}
\label{eq1}
    PET = t_2-t_1,
\end{equation}

\begin{figure}
    \centering
    \includegraphics[scale=0.33]{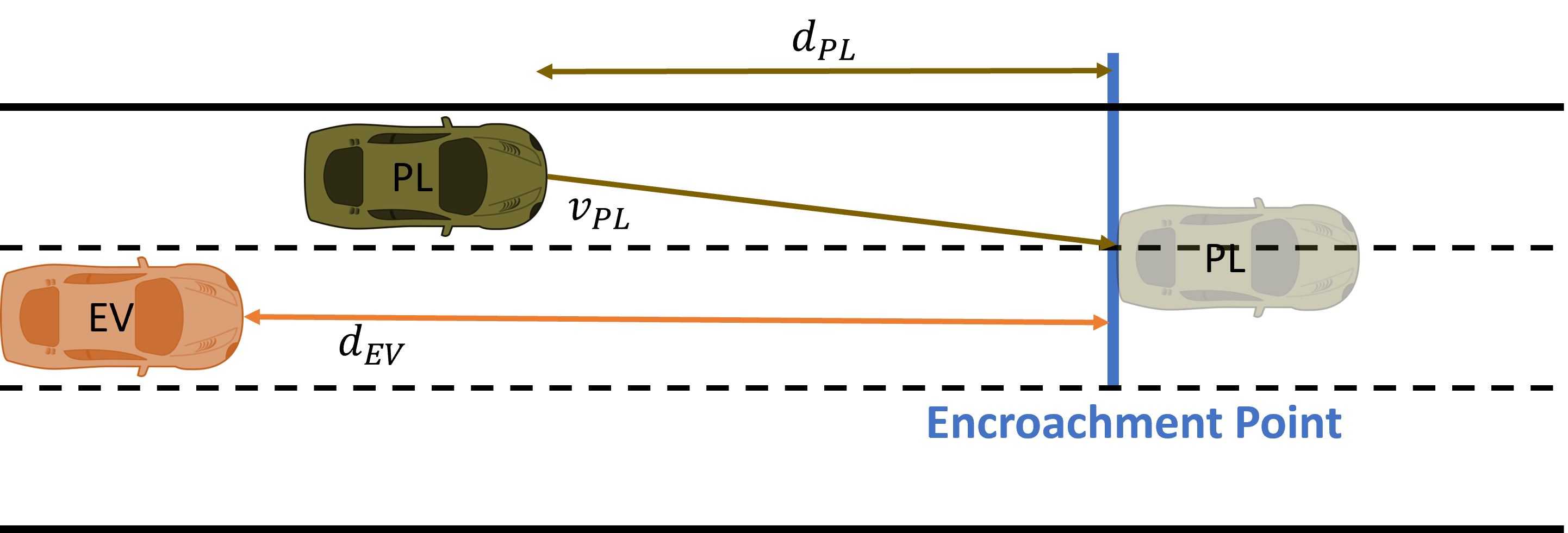}
    \caption{The computation of Post Encroachment Time (PET) for parallel-moving vehicles involves assessing the lateral velocity component of $v_{PL}$. If this component signals a potential encroachment, defined by more than half of the width of the parallel vehicle crossing into the lane of the ego vehicle (EV) at the encroachment point, the PET can be estimated.}
    \label{fig:PET}
\end{figure}

Given that our specific scenario did not encompass an intersection, our calculation for PET of parallel vehicles involved determining whether their trajectories would intersect with the lane of the ego vehicle, assuming a constant velocity. The moment of potential intersection was defined as the moment when the parallel vehicle would enter with more than half its width in the lane of the EV. This defined a designated encroachment point for our analysis. Parallel vehicles would further be classified as parallel-leading (PL) if the encroachment point is reached in front of the EV and as parallel-following (PF) if it reached after the EV. Figure~\eqref{fig:PET} illustrates the PET calculation.\\
For a vehicle already in the lane of the EV the PET calculation simplifies since the encroachment point is the current position of the leading vehicle. Therefore, with $t_1=0$, PET is equivalent to the time required for the following vehicle to reach the leading vehicle, which corresponds to the definition of TH in equation ~\eqref{eq2}. 
\begin{equation}
\label{eq2}
    TH = \frac{d}{v} = t_2,
\end{equation}

where $d$ represents the distance between the two vehicles, $v$ denotes the velocity of the following vehicle, and $t_2$ is the time for the following vehicle to reach the encroachment point as defined in \eqref{eq1}. Therefore, utilizing PET in car following scenarios is equivalent to utilizing TH. Consequently, TH was not employed in addition to PET.\\

To investigate car-following scenarios, the conventional TTC lacks a precise definition for non-collision trajectories. In addressing this limitation, we adopt in this work an alternative approach by utilizing the inverse, introduced in ~\cite{ITTC}, as denoted in equation~\eqref{eq3},

\begin{equation}
\label{eq3}
    ITTC = \frac{v_F-v_L}{d},
\end{equation}

where $v_{L}, v_F$ represents the velocity of the leading and following vehicles and $d$ denotes the distance between them.  
%In situations where the PET calculation indicates a potential collision, resulting in zero distance between the two vehicles, the predicted collision time is used for the ITTC. 
If the two vehicles are not on a collision course, ITTC is set to $\infty$.\\

In contrast to time-based SSMs, deceleration-based SSMs such as DRAC assess collision risk by evaluating a vehicle's capability to decelerate adequately and avert a crash. When two vehicles are not on a collision trajectory, the DRAC value is set to 0, indicating the absence of a required deceleration for collision prevention as denoted in equation ~\eqref{eq4},

\begin{equation}
\label{eq4}
    DRAC = \begin{cases}
    \frac{(v_F-v_L)^2}{d}, & v_F > v_L \\
    0, & else
    \end{cases}
\end{equation}

where $v_{F}$ and $v_{L}$ represent the velocity of the following and leading vehicle respectively and $d$ denotes the distance between them.

\subsection{Safety risk estimation}
\label{sec:Framework}

The objective of the safety model framework is to establish a connection between objective safety measures, i.e. SSMs, and subjective driving behavior. The safety assessment is thus split into two parts. Initially, the safety risk of the traffic situation between the ego vehicle and each surrounding traffic participant is assessed using SSMs. Subsequently, the overall safety risk for the ego vehicle is calculated by aggregating the calculated safety risks.

The safety risk due to a single vehicle is obtained from the SSM values of the traffic situation of between surrounding vehicles and the ego vehicle. Those values are combined into a safety risk value through two methods:
\begin{itemize}
    \item Grid-based combination: The individual SSM values are used to categorize the traffic situation into \textit{safe}, \textit{conflict} and \textit{critical} categories, with the respective thresholds in Table \ref{tbl:SSM_thresholds} taken from literature and not recalibrated for the HighD dataset. Additional insights into this methodology can be found in \cite{9742814}.
    \item AutoEncoder based MeanAbsoluteError (MAE): An AutoEncoder (AE) is trained on the \textit{safe} trajectories (according to all three SSMs) utilizing normalized SSM values via hyperbolic tangent functions. The MAE is used as a safety risk estimation, in a similar procedure to \cite{EGK_AE}. The main advantage of this method is the direct use of SSM values instead of categories and thus smoother safety risk evaluations.
\end{itemize}

The overall safety risk for the ego vehicle results from a linear combination of the risk associated with the surrounding vehicles. In order to address the influence of relative positioning on subjective safety (\cite{9987124} and \cite{10186747}), we weight the contribution of the individual safety risk depending on the vehicle's position:

\begin{itemize}
    \item LV: Leading Vehicle situated in the same lane
    \item FV: Following Vehicle situated in the same lane
%    \item PC: Parallel Vehicle on a collision course
    \item PL: Parallel Vehicle anticipated to merge ahead of the ego vehicle.
    \item PF: A vehicle anticipated to merge behind the ego vehicle.
\end{itemize}

\begin{table}
\caption{Safety categories are determined based on predefined safety thresholds corresponding to each SSM, according to  \cite{UN157}, \cite{SSM}, and \cite{9742814}. 
Each safety category is assigned a numerical value for further quantification and analysis.}
\centering
\begin{tabular}{l|lll}
\multirow{2}{*}{SSM} & \multicolumn{3}{c}{Safety category} \\  
                     & Safe    & Conflict    & Critical    \\ \hline
TH/PET               & $[1,\infty]$&$[0.4, 1]$ & $[0, 0.4]$  \\
DRAC                 & $[0, 3.3]$  &    $[5, 3.3]$         & $[5, \infty]$            \\
ITTC                 & $[-\infty, 1/1.5]$ & $[1/1.5, 1]$& $[1,\infty]$     \\
\hline
Safety Risk & 0 & 0.5 & 1
\end{tabular}
\label{tbl:SSM_thresholds}
\end{table}

\section{Model evaluation}
\label{evaluation}

The safety model generates a time series depicting the safety risk for each ego vehicle. The correlation between the gradient of the safety risk and the jerk serves as an indicator of the model's precision in assessing the safety risk as perceived by the human driver.

As the model takes into account simultaneous changes in the safety risk arising from both leading and following vehicles, the corresponding driving behavior-related reaction can involve either deceleration or acceleration. Consequently, we examined the correlation between the absolute values of both time series.

A time delay between the safety gradient and jerk (due to the reaction time) was expected. Therefore, we first determined the time lag corresponding to the maximum cross-correlation. If this time delay was within the range of $[0,2]$ seconds, implying a potential response to the safety risk change within a reasonable reaction time, the two time series were shifted accordingly. Subsequently, the Spearman rank correlation test was employed to assess the monotonic correlation between the two variables, considering the absence of a strictly linear relationship.\\

In a first step we evaluated the correlation between the single vehicle's safety risk and the ego vehicle's driving behavior for different grid-based combinations and AE shown in Table \ref{tab:SSM_weights}. In the first grid configuration (a) an equal weight is given to all SSMs, in configuration (b) the influence of PET/TH is increased and configurations (c),(d), and (e) only utilize a single SSM. Configurations (f) and (g) utilize two variations of the AE instead, one with a linear output layer and MAE loss function and one using tanh activation function and Binary cross-entropy loss.

%The examined SSM weights are detailed in Table \ref{tab:SSM_weights}. In the first configuration (a), distance-based SSMs had twice the weight of collision-based SSMs, with each type contributing equally to the total weight. Configuration (b) placed double weight on collision-based SSMs, while the final configuration (f) distributed an equal weight to all SSMs. Configurations (c), (d), and (e) were utilized to assess the impact of using individual SSMs rather than combinations.

The second step combines the SSM configurations with positional weight configurations listed in Table \ref{tab:pos_weights}. The first configuration focuses only on the leading and following vehicles, The second configuration considers the impact on the overall safety risk from all surrounding vehicles equally. The final configuration (3) utilizes the observations from the single vehicles safety risk correlation analysis and emphasizes the surrounding vehicles.

In addition to comparing the means of the correlation coefficients, the distributions were assessed using the non-parametric Wilcoxon ranked sign test with the null hypothesis proposing that the difference between pairs from the distributions of different models is centered around 0. The alternative hypothesis was defined such that the center is $> 0$, indicating that one model produced a higher correlation across the overall dataset.

\begin{table}
\caption{Weights and AE model used to calculate the individual safety risk of each surrounding vehicle.}
\label{tab:SSM_weights}
\centering
\begin{tabular}{l||lll} 
\#Configuration & PET/TH & DRAC & ITTC \\ \hline \hline 
a                & $1/3$      & $1/3$    & $1/3$    \\ \hline 
b                & $2/3$      & $1/6$    & $1/6$    \\ \hline 
c                & 1      & -    & -    \\ \hline 
d                & -      & 1    & -    \\ \hline 
e                & -      & -    & 1    \\ \hline 
f                & \multicolumn{3}{c}{AE linear} \\ \hline
g                & \multicolumn{3}{c}{AE tanh} \\
\end{tabular}
\end{table}

\begin{table}
\caption{Weights assigned based on the position of other vehicles to estimate the contribution to the safety of the ego vehicle. L represents the leading vehicle, F denotes the following vehicle, and PL/PF refers to parallel vehicles whose trajectories are estimated to cross into the lane of the ego vehicle in front/behind it.}
\centering
\begin{tabular}{l||llll}
\multicolumn{1}{l||}{\multirow{2}{*}{Configuration}} & \multicolumn{4}{c}{relative position}     \\
\multicolumn{1}{l||}{}                                  & L   & F    & PL   & PF    \\ \hline \hline 
1                                                      & 1   & 1  & -    & -     \\ \hline 
2                                                      & 1   & 1       & 1    & 1     \\ \hline 
3                                                      & 1   & 1   & 2 & 2
\end{tabular}
\label{tab:pos_weights}
\end{table}

\section{Model Evaluation Results}
\label{sec:Results}

\begin{figure}
    \centering
    \includegraphics[scale=0.5]{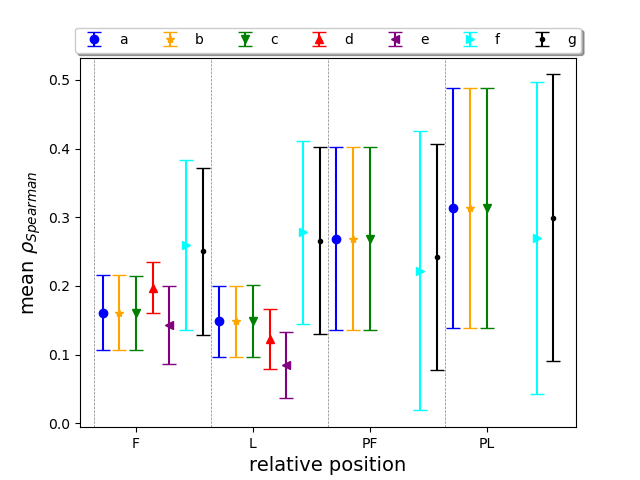}
    \caption{Means and standard deviations of the statistically significant Spearman coefficients for various SSM weights and AEs from Table \ref{tab:SSM_weights}. Weight combinations without PET are unable to evaluate parallel vehicles.}
    \label{fig:spearman_pos}
\end{figure}

Within the HighD dataset the safety risk for 70040 ego vehicles was analyzed, with a total of ca. 19 mio. L and F vehicles, 12 mio. PL and 14 mio. PF vehicles. Each individual vehicle can appear one as ego vehicle, but multiple times as part of the surrounding vehicles.
In a first step, the influence of different SSM weights on the correlation for vehicles in different relative positions was analyzed, with the results shown in Figure \ref{fig:spearman_pos}.

As per definition, SSM weight configurations without PET (d-e) are unable to assess both PL and PF, however, they perform fairly well for assessing the following vehicle. The AE-based assessments show overall the highest mean correlation, and also the highest standard deviation. 
The fraction of significant and non-significant correlations is shown in Table  \ref{tab:ratio_pos} and generally follows the same trend shown in Figure \ref{fig:spearman_pos}.

\begin{table}
\caption{Fraction of significant/non-significant correlations of the single vehicle safety assessment for various SSM weights and AEs from Table \ref{tab:SSM_weights}}
\label{tab:ratio_pos}
\centering
\begin{tabular}{l||llll}
\multicolumn{1}{l||}{\multirow{2}{*}{Configuration}} & \multicolumn{4}{c}{relative position}\\
 & F & L & PF & PL\\ \hline \hline 
a& 0.025164      & 0.022498    & 0.219520 &   0.166757  \\ \hline 
b& 0.025208     & 0.022498   & 0.219520 & 0.166757   \\ \hline 
c& 0.025106      & 0.022367    & 0.218976  & 0.166408  \\ \hline 
d& 0.000363      & 0.000087    & 0 & 0    \\ \hline 
e& 0.000174      & 0.000073    & 0 & 0   \\ \hline 
f& 0.240178 & 0.423934 & 0.152531 & 0.151106 \\ \hline
g&  0.230787 & 0.367319 & 0.158463 & 0.161871 \\
\end{tabular}
\end{table}

Based on these results we decided to only utilize the configurations (a),(b),(f) and (g) to assess the overall safety risk for all ego vehicles. Though (a) and (b) perform similarly, we utilize both to analyze any difference from varying the weights. Both AE-based configurations are utilized for the same reason.
Configurations (d) and (e) ignore all surrounding vehicles, whereas configuration (c) does not show any meaningful difference from configurations (a) and (b).

\begin{figure}
    \centering
    \includegraphics[scale=0.5]{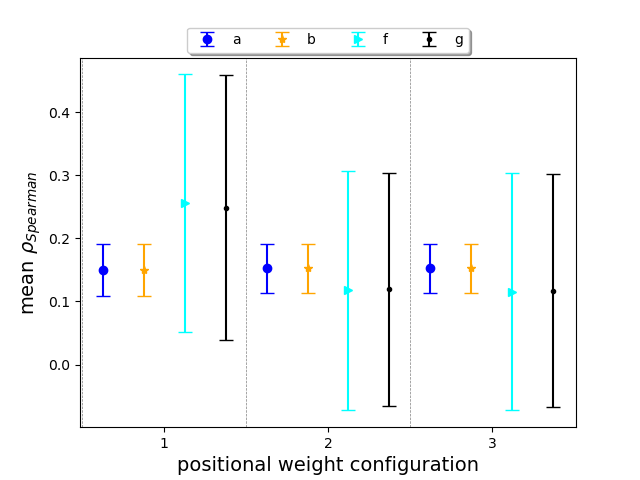}
    \caption{ Means and standard deviations of the Spearman coefficient for various model configurations $1a3g$ on the HighD dataset, indicating the correlation between jerk and the gradient of the safety risk.}
    \label{fig:spearman}
\end{figure}

The distributions of the statistically significant ($p_{\text{value}} < 0.05$) Spearman correlation coefficient ($\rho$) for the 12 resulting models $1a-3g$ is illustrated in Figure \ref{fig:spearman}.
The figure is categorized by both positional weights (x-axis) and SSM weights (marker color and shape). 

Similarly to Figure \ref{fig:spearman_pos}, models using SSM configurations (f) and (g) have the highest mean correlation coefficient when only the leading and following vehicles are considered, i.e. models (1a,1b,1f,1g), though they still show a high standard deviation. When considering the parallel vehicles (2a-3g), the grid-based SSM configurations perform better. Their mean correlation coefficient stays near constant also when considering only leading and following vehicles (1a,1b). However, the fraction of significant/non-significant correlations does improve dramatically for positional weights 2 and 3, as shown in Table \ref{tab:ratio_full}. The highest fraction of significant correlations is found for the models (1f) and (1g).

\begin{table}
\caption{Fraction of significant/non-significant correlations of the overall ego vehicle safety assessment for various SSM weights and AEs from Table \ref{tab:SSM_weights} and positional weights from Table \ref{tab:pos_weights}.}
\label{tab:ratio_full}
\centering
\begin{tabular}{l||lll}
\multicolumn{1}{l||}{SSM weight} & \multicolumn{3}{c}{Positional weight configuration}\\
 configuration & 1 & 2 & 3 \\ \hline \hline 
a& 0.094838      & 0.246177	   & 0.236253   \\ \hline 
b& 0.092882     & 0.242450   & 0.236653    \\ \hline 
f& 0.308733 & 0.177916 & 0.174832  \\ \hline
g&  0.302062 & 0.179175 & 0.176305 \\
\end{tabular}
\end{table}

For a more comprehensive comparison, we utilized the Wilcoxon rank test to compare the distributions of the significant correlation coefficients. The results are shown in Table \ref{tab:Wil}, with the null hypothesis defined as a symmetric distribution of the rank differences of the coefficients around 0 and the alternative as a positive shift of the differences. The rank differences are calculated as $row-column$, e.g. the entry in the row (1f) and column (1a) indicates that the Null hypothesis was rejected and the rank differences of (1f-1a) are shifted positively.

Overall, the Wilcoxon rank test shows the models (1f) and (1g) as the best-performing ones, though (*f) and (*g) models perform worst for other positional weights. Model (2a) performs better than (2b), yet (3b) performs better than (3a).

\begin{table}
\caption{Results of the pairwise Wilcoxon rank test for the Spearman correlation coefficients. The alternative hypothesis to the null hypothesis of a symmetric distribution around zero is set as the distribution being shifted positively. An entry of "\textbf{r}" signifies that the Null hypothesis was rejected with a p-value $< 0.05$ and the correlation coefficient distribution from the model configuration defined by the row is shifted positively compared to the one defined by the column. An entry of "n" signifies that the Null hypothesis was not rejected.}

\centering
\setlength\tabcolsep{4pt}
\begin{tabular}{l||llllllllllll}
   & 1a         & 1b         & 1f & 1g         & 2a         & 2b         & 2f         & 2g         & 3a         & 3b         & 3f         & 3g         \\ \hline \hline
1a & -          & n          & n  & n          & n          & n          & n          & n          & n          & n          & n          & n          \\
1b & n          & -          & n  & n          & n          & n          & n          & n          & n          & n          & n          & n          \\
\textbf{1f}& \textbf{r} & \textbf{r} & -  & \textbf{r} & \textbf{r} & \textbf{r} & \textbf{r} & \textbf{r} & \textbf{r} & \textbf{r} & \textbf{r} & \textbf{r} \\
1g & \textbf{r} & \textbf{r} & n  & -          & \textbf{r} & \textbf{r} & \textbf{r} & \textbf{r} & \textbf{r} & \textbf{r} & \textbf{r} & \textbf{r} \\
2a & \textbf{r} & \textbf{r} & n  & n          & -          & \textbf{r} & \textbf{r} & \textbf{r} & \textbf{r} & \textbf{r} & \textbf{r} & \textbf{r} \\
2b & \textbf{r} & \textbf{r} & n  & n          & n          & -          & \textbf{r} & \textbf{r} & \textbf{r} & \textbf{r} & \textbf{r} & \textbf{r} \\
2f & \textbf{r} & \textbf{r} & n  & n          & n          & n          & -          & n          & n          & n          & \textbf{r} & n          \\
2g & \textbf{r} & \textbf{r} & n  & n          & n          & n          & n          & -          & n          & n          & \textbf{r} & \textbf{r} \\
3a & \textbf{r} & \textbf{r} & n  & n          & n          & n          & \textbf{r} & \textbf{r} & -          & n          & \textbf{r} & \textbf{r} \\
3b & \textbf{r} & \textbf{r} & n  & n          & n          & n          & \textbf{r} & \textbf{r} & \textbf{r} & -          & \textbf{r} & \textbf{r} \\
3f & \textbf{r} & \textbf{r} & n  & n          & n          & n          & n          & n          & n          & n          & -          & n          \\
3g & \textbf{r} & \textbf{r} & n  & n          & n          & n          & n          & n          & n          & n          & \textbf{r} & -         
\end{tabular}
\label{tab:Wil}
\end{table}

Besides the quantitative analysis with the HighD dataset we tested one model on a highway traffic scenario from the IAMCV dataset, since we could observe the selected scene for a longer duration and ensure no interference due to past or future traffic.

\begin{figure}
    \centering
    \includegraphics[scale=0.3]{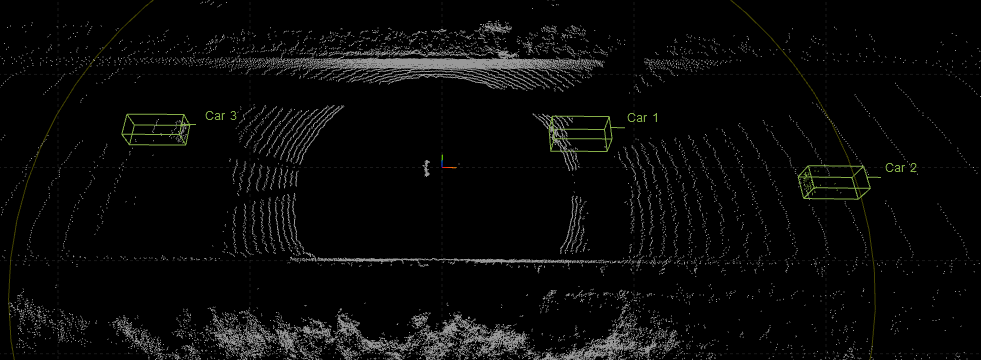}
    \caption{Car following scenario on the highway from the IAMCV dataset.}
    \label{fig:IAMCV}
\end{figure}

\begin{figure}
    \centering
    \includegraphics[scale=0.5]{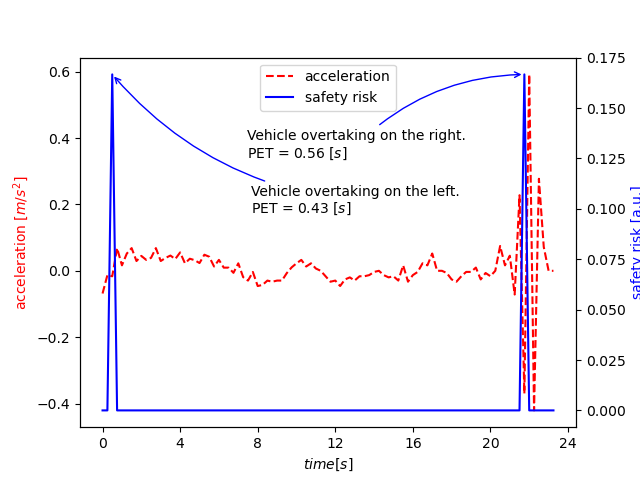}
    \caption{Timeseries of acceleration and safety risk for a single vehicle with the model configuration 2a. Two minor and short traffic conflicts occur in the recording, with the second one leading to a reaction by the ego vehicle.}
    \label{fig:model_results}
\end{figure}

The timeseries for the safety risk of the traffic scenario extracted from the IAMCV dataset, depicted in Figure \ref{fig:IAMCV}, is shown in Figure \ref{fig:model_results}.
The ego vehicle maintained a substantial distance ($2s$) from the car ahead, ensuring minimal safety risk for the majority of the recording. However, there is a slight elevation in the estimated safety risk near the beginning, due to a car overtaking on the left and slightly invading the ego vehicle's lane (refer to Figure \ref{fig:PET}). Extrapolating the trajectories linearly generates a PET of $0.43 [s]$ and an overall safety risk of $0.16$.

A comparable incident occurs towards the recording's end, with a car overtaking on the right. In this instance, the ego vehicle exhibits a pronounced reaction, evidenced by a statistically significant Spearman coefficient $\rho = 0.34$ and a time lag of $0.25 [s]$.

\section{Conclusion and Future Work}
\label{sec:Conclusion}
In this study, we introduced a framework for simultaneously employing multiple SSMs in a multi-vehicle scenario to assess safety, demonstrated through its application in a highway car-following situation. 
We explored both the effect of merging the SSMs of individual vehicles using a grid-like structure and utilizing Autoencoders to obtain a safety risk value. The method based on autoencoders demonstrated a  strong performance; however, it exhibited significant variance in its correlation coefficients, suggesting a less stable model. %The latter showed a surprisingly good performance, though suffer from a large variance in their correlation coefficients which indicates a less stable model.
The incorporation of all surrounding vehicles showed a significant improvement for grid-based SSM combinations, resulting in the model with the best stability and comparatively high correlation frequency. However, variations in positional weights did not significantly improve the models.
In general, the significant correlation coefficients indicate a moderate correlation between the safety risk gradient and jerk for all models. This is partially explained by the potential impact of non-safety-related considerations on ego vehicle behavior. However, though more model configurations were tested than shown in this publication, no dedicated optimization has yet been done beyond a grid-search for model configurations. This, and further refinement by adjusting SSM thresholds based on velocity, similar to those outlined in UN 157, or by utilizing additional SSMs and improving the AEs offer opportunities for adaptation across diverse domains without necessitating alterations to the underlying framework. The framework can further enable the simulation of human driving behavior in dangerous situations, for example, involving hydrogen-powered vehicles that could rupture in a collision, releasing hydrogen gas and potentially leading to an explosion, where real data is still lacking.

We conclude that the framework presented in this work enables the integration of insights in subjective perception of safety from naturalistic data, while utilizing objective metrics. Thus it serves as a foundation for developing a SSM-driven safety field model, combining the strengths of both approaches and adopting a 2-dimensional perspective of the scenario with SSMs and multiple vehicles.



\section*{ACKNOWLEDGMENT}

This work was supported by the Austrian Science Fund (FWF), project number P 34485-N and by Johannes Kepler University Linz, Linz Institute of Technology (LIT), project number LIFTC-2022-1-117, funded by the State of Upper Austria and the Federal Ministry of Education, Science and Research. 

%%%%%%%%%%%%%%%%%%%%%%%%%%%%%%%%%%%%%%%%%%%%%%%%%%%%%%%%%%%%%%%%%%%%%%%%%%%%%%%%

\bibliographystyle{IEEEtran}
\bibliography{bliblio}

\begin{thebibliography}{10}
\providecommand{\url}[1]{#1}
\csname url@samestyle\endcsname
\providecommand{\newblock}{\relax}
\providecommand{\bibinfo}[2]{#2}
\providecommand{\BIBentrySTDinterwordspacing}{\spaceskip=0pt\relax}
\providecommand{\BIBentryALTinterwordstretchfactor}{4}
\providecommand{\BIBentryALTinterwordspacing}{\spaceskip=\fontdimen2\font plus
\BIBentryALTinterwordstretchfactor\fontdimen3\font minus \fontdimen4\font\relax}
\providecommand{\BIBforeignlanguage}[2]{{%
\expandafter\ifx\csname l@#1\endcsname\relax
\typeout{** WARNING: IEEEtran.bst: No hyphenation pattern has been}%
\typeout{** loaded for the language `#1'. Using the pattern for}%
\typeout{** the default language instead.}%
\else
\language=\csname l@#1\endcsname
\fi
#2}}
\providecommand{\BIBdecl}{\relax}
\BIBdecl

\bibitem{ADAS_impact}
\BIBentryALTinterwordspacing
L.~Masello, G.~Castignani, B.~Sheehan, F.~Murphy, and K.~McDonnell, ``On the road safety benefits of advanced driver assistance systems in different driving contexts,'' \emph{Transportation Research Interdisciplinary Perspectives}, vol.~15, p. 100670, 2022. [Online]. Available: \url{https://www.sciencedirect.com/science/article/pii/S2590198222001300}
\BIBentrySTDinterwordspacing

\bibitem{european2019eu}
\emph{EU Road Safety Policy Framework 2021--2030 - Next Steps Towards “Vision Zero”}.\hskip 1em plus 0.5em minus 0.4em\relax European Commission Brussels, 2019.

\bibitem{UN157}
``Un regulation no 157 – uniform provisions concerning the approval of vehicles with regards to automated lane keeping systems [2021/389],'' pp. 75--137, Mar 2021.

\bibitem{DSSM}
S.~Tak, S.~Kim, D.~Lee, and H.~Yeo, ``A comparison analysis of surrogate safety measures with car-following perspectives for advanced driver assistance system,'' \emph{Journal of Advanced Transportation}, vol. 2018, 11 2018.

\bibitem{10186747}
E.~Del~Re and C.~Olaverri-Monreal, ``Method for comparison of surrogate safety measures in multi-vehicle scenarios,'' in \emph{2023 IEEE Intelligent Vehicles Symposium (IV)}, 2023, pp. 1--6.

\bibitem{9987124}
E.~del Re and C.~Olaverri-Monreal, ``Implementation of road safety perception in autonomous vehicles in a lane change scenario,'' in \emph{2022 IEEE International Conference on Vehicular Electronics and Safety (ICVES)}, 2022, pp. 1--6.

\bibitem{highD}
R.~Krajewski, J.~Bock, L.~Kloeker, and L.~Eckstein, ``The highd dataset: A drone dataset of naturalistic vehicle trajectories on german highways for validation of highly automated driving systems,'' in \emph{2018 IEEE 21st International Conference on Intelligent Transportation Systems (ITSC)}, 2018.

\bibitem{icmcv}
N.~Certad, E.~Del~Re, H.~Korndoerfer, G.~Schroedinger, W.~Morales-Alvarez, S.~Tschernuth, D.~Gankhuyag, L.~Del~Re, and C.~Olaverri-Monreal, ``{IAMCV Mulit-Scenario Vehicle Interaction Dataset},'' \emph{{IEEE Intelligent Transportation Systems Magazine}}, 2024.

\bibitem{certad2023}
\BIBentryALTinterwordspacing
N.~Certad, E.~Del~Re, A.~Aghanouri, D.~Gankhuyag, W.~Morales-Alvarez, S.~Tschernuth, L.~Del~Re, and C.~Olaverri-Monreal. (2023) {IAMCV Interaction of Autonomous and Manually Controlled Vehicles}. {IEEE Dataport}. [Online]. Available: \url{https://dx.doi.org/10.21227/d1g3-c160}
\BIBentrySTDinterwordspacing

\bibitem{SSM}
S.~M. Mahmud, L.~Ferreira, M.~Hoque, and A.~Hojati, ``Application of proximal surrogate indicators for safety evaluation: A review of recent developments and research needs,'' \emph{IATSS Research}, vol.~41, 03 2017.

\bibitem{SSM_2023}
T.~Das, M.~S. Samandar, M.~K. Autry, and N.~M. Rouphail, ``Surrogate safety measures: Review and assessment in real-world mixed traditional and autonomous vehicle platoons,'' \emph{IEEE Access}, vol.~11, pp. 32\,682--32\,696, 2023.

\bibitem{MTTC}
K.~Ozbay, P.~Associate, B.~Bartin, and H.~Yang, ``Derivation and validation of new simulation-based surrogate safety measure,'' \emph{Transportation Research Record}, vol. 2083, 12 2008.

\bibitem{fuzzy}
\BIBentryALTinterwordspacing
K.~Mattas, M.~Makridis, G.~Botzoris, A.~Kriston, F.~Minarini, B.~Papadopoulos, F.~Re, G.~Rognelund, and B.~Ciuffo, ``Fuzzy surrogate safety metrics for real-time assessment of rear-end collision risk. a study based on empirical observations,'' \emph{Accident Analysis \& Prevention}, vol. 148, p. 105794, 2020. [Online]. Available: \url{https://www.sciencedirect.com/science/article/pii/S0001457520316146}
\BIBentrySTDinterwordspacing

\bibitem{SSAM}
\BIBentryALTinterwordspacing
L.~Vasconcelos, L.~Neto, Álvaro Maia~Seco, and A.~B. Silva, ``Validation of the surrogate safety assessment model for assessment of intersection safety,'' \emph{Transportation Research Record}, vol. 2432, no.~1, pp. 1--9, 2014. [Online]. Available: \url{https://doi.org/10.3141/2432-01}
\BIBentrySTDinterwordspacing

\bibitem{Combined_SSM}
\BIBentryALTinterwordspacing
A.~Mazaheri, M.~Saffarzadeh, N.~Nadimi, and S.~S. Naseralavi, ``A revise on using surrogate safety measures for rear-end crashes,'' \emph{IATSS Research}, vol.~47, no.~1, pp. 105--120, 2023. [Online]. Available: \url{https://www.sciencedirect.com/science/article/pii/S0386111223000092}
\BIBentrySTDinterwordspacing

\bibitem{ESRA}
G.~Furian, S.~Kaiser, N.~Senitschnig, and A.~Soteropoulos, ``Subjective safety and risk perception,'' 2021.

\bibitem{TTC_perception}
A.~Martín, A.~Décima, and J.~Barraza, ``Perception of speed, distance, and ttc of familiar objects,'' \emph{Psychology \& Neuroscience}, vol.~10, pp. 261--272, 09 2017.

\bibitem{mirror}
K.~Shimono and A.~Higashiyama, ``Perceived size and distance of virtual targets in convex mirrors,'' \emph{Perception}, vol.~29, pp. 33--33, 01 2000.

\bibitem{mirror_adjustment}
\BIBentryALTinterwordspacing
C.~Böffel and J.~Müsseler, ``Adjust your view! wing-mirror settings influence distance estimations and lane-change decisions,'' \emph{Transportation Research Part F: Traffic Psychology and Behaviour}, vol.~35, pp. 112--118, 2015. [Online]. Available: \url{https://www.sciencedirect.com/science/article/pii/S1369847815001606}
\BIBentrySTDinterwordspacing

\bibitem{safety_field_infra}
F.~A. Mullakkal-Babu, M.~Wang, X.~He, B.~{van Arem}, and R.~Happee, ``Probabilistic field approach for motorway driving risk assessment,'' \emph{Transportation Research Part C Emerging Technologies}, vol. 118, 07 2020.

\bibitem{physics_field}
A.~Arun, S.~M.~M. Haque, S.~Washington, and F.~Mannering, ``A physics-informed road user safety field theory for traffic safety,'' 10 2022.

\bibitem{SSM_model_india}
P.~Diwakar, V.~Landge, and U.~Jain, ``Evaluating the relationship between surrogate safety measures and traffic event severity in terms of human perception of danger: A perspective under indian traffic conditions,'' \emph{Applied Sciences}, vol.~13, p. 12100, 11 2023.

\bibitem{ITTC}
V.~E. Balas and M.~M. Balas, ``Driver assisting by inverse time to collision,'' \emph{2006 World Automation Congress}, pp. 1--6, 2006.

\bibitem{9742814}
E.~Del~Re and P.~Tkachenko, ``A grid-based surrogate safety measure for traffic safety assessment,'' in \emph{2022 International Conference on Connected Vehicle and Expo (ICCVE)}, 2022, pp. 1--6.

\bibitem{EGK_AE}
N.~Arslan, D.~Özdemir, and H.~Temurtas, ``Ecg heartbeats classification with dilated convolutional autoencoder,'' \emph{Signal, Image and Video Processing}, vol.~18, pp. 1--10, 09 2023.

\end{thebibliography}

\end{document}